\documentclass[conference]{IEEEtran}
\IEEEoverridecommandlockouts

\usepackage{cite}
\usepackage{amsmath,amssymb,amsfonts}
\usepackage{algorithmic}
\usepackage{graphicx}
\usepackage{textcomp}
\usepackage{xcolor}
\usepackage{multirow}
\def\BibTeX{{\rm B\kern-.05em{\sc i\kern-.025em b}\kern-.08em
    T\kern-.1667em\lower.7ex\hbox{E}\kern-.125emX}}
\begin{document}

\title{NeuCODEX: Edge–Cloud Co-Inference with Spike-Driven Compression and Dynamic Early-Exit\\
}

\author{\IEEEauthorblockN{Maruf Hassan}
\IEEEauthorblockA{\textit{Centre for Sustainable Digital Technologies} \\
\textit{Technological University Dublin}\\
Dublin, Ireland \\
maruf.hassan@tudublin.ie}
\and
\IEEEauthorblockN{Steven Davy}
\IEEEauthorblockA{\textit{Centre for Sustainable Digital Technologies} \\
\textit{Technological University Dublin}\\
Dublin, Ireland \\
steven.davy@tudublin.ie}
\and
\IEEEauthorblockN{Muhammad Zawish}
\IEEEauthorblockA{\textit{Centre for Sustainable Digital Technologies} \\
\textit{Technological University Dublin}\\
Dublin, Ireland \\
muhammad.zawish@tudublin.ie}
\and
\IEEEauthorblockN{Owais Bin Zuber}
\IEEEauthorblockA{\textit{Smart Network Innovation Lab}\\
\textit{Huawei Ireland Research Center}\\
Dublin, Ireland \\
owais.bin.zuber@huawei.com}
\and
\IEEEauthorblockN{Nouman Ashraf}
\IEEEauthorblockA{\textit{Centre for Sustainable Digital Technologies} \\
\textit{Technological University Dublin}\\
Dublin, Ireland \\
nouman.ashraf@tudublin.ie}
}

\maketitle

\begin{abstract}
Spiking Neural Networks (SNNs) offer significant potential for enabling energy-efficient intelligence at the edge. However, performing full SNN inference at the edge can be challenging due to the latency and energy constraints arising from fixed and high timestep overheads. Edge-cloud co-inference systems present a promising solution, but their deployment is often hindered by high latency and feature transmission costs. To address these issues, we introduce NeuCODEX, a neuromorphic co-inference architecture that jointly optimizes both spatial and temporal redundancy. NeuCODEX incorporates a learned spike-driven compression module to reduce data transmission and employs a dynamic early-exit mechanism to adaptively terminate inference based on output confidence. We evaluated NeuCODEX on both static images (CIFAR10 and Caltech) and neuromorphic event streams (CIFAR10-DVS and N-Caltech). To demonstrate practicality, we prototyped NeuCODEX on ResNet-18 and VGG-16 backbones in a real edge-to-cloud testbed. Our proposed system reduces data transfer by up to 2048x and edge energy consumption by over 90\%, while reducing end-to-end latency by up to $3\times$ compared to edge-only inference, all with a negligible accuracy drop of less than 2\%. In doing so, NeuCODEX enables practical, high-performance SNN deployment in resource-constrained environments.
\end{abstract}


\section{Introduction}
Recent progress in neuromorphic and edge artificial intelligence (AI) systems has led to a growing interest in utilizing spiking neural networks (SNNs) for energy-efficient and low-latency edge applications. SNNs involve time-dependent processing of information using discrete spike events. This temporal processing inherently produces sparse activations, which contribute to computational efficiency. By substituting expensive multiply-accumulate (MAC) operations with straightforward accumulations, SNNs achieve low-power computation \cite{roy2019towards}. However, deploying deep and complex SNNs entirely on low-power edge devices can still be challenging due to limited computation and energy resources. Partial offloading to the cloud presents a promising solution; however, it naturally involves the overhead of transferring intermediate features over bandwidth-limited networks. 

Prior works on edge–cloud co-inference (also known as split computing or collaborative intelligence) have shown that intelligently inserting bottleneck compression units at the split point of a neural network to reduce the size of the data transmitted can significantly improve latency and energy consumption \cite{shao2020bottlenet++}. This lossy and spatial feature compression can drastically cut bandwidth requirements in split inference systems \cite{hassan2024spikebottlenet},\cite{wang2024snn}.

Beyond spatial compression, another dimension of efficiency in SNNs is the temporal depth processing. Unlike traditional networks that produce an output in a single forward pass, SNNs accumulate information across multiple timesteps. While more timesteps can improve accuracy, many inputs can be classified correctly with fewer steps \cite{li2023seenn}. This observation led to dynamic early-exit mechanisms that terminate inference once a confident prediction is reached, saving computation and energy.

Despite progress in spike-driven feature compression and dynamic early exits separately, they have largely been studied in isolation. We identify a crucial opportunity to combine these techniques in an edge–cloud SNN co-inference system. Compressing spike communications reduces bandwidth, while skipping redundant timesteps minimizes latency and energy, jointly addressing communication and compute inefficiencies in neuromorphic deployment.

In this work, we present NeuCODEX, \textbf{Neu}romorphic  \textbf{Co}-Inference Architecture with Spike-Driven Compression and \textbf{D}ynamic \textbf{E}arly-\textbf{Ex}it Mechanism, a unified framework that integrates a spike-driven bottleneck encoder–decoder with a confidence-based early-exit mechanism for SNN co-inference. NeuCODEX's novelty lies in the architectural synergy of jointly exploiting spatial compression and temporal adaptation in a neuromorphic split computing scenario. The goal is to enable efficient edge–cloud SNN inference that is communication-aware and runtime-adaptive, which is crucial for applications like internet of things (IoT) sensors or autonomous systems where both bandwidth and energy are at a premium.

In summary, our contributions are: 

\begin{itemize}
    \item A novel neuromorphic co-inference architecture that integrates a learned spike compression bottleneck and a dynamic timestep early-exit mechanism.
    
    \item An implementation on two network backbones, ResNet-18 and VGG-16, and evaluation on both static image datasets (CIFAR-10, Caltech) and neuromorphic event stream datasets (CIFAR10-DVS, NCaltech) in a real edge-to-cloud deployment.
    
    \item Demonstration of significant performance gains, including up to 2048$\times$ reduction in transmitted bits and over 90\% reduction in edge energy, with 31--37\% lower latency versus edge-only execution and up to 97\% latency reduction compared to naive uncompressed offloading, all with \textasciitilde 1-2\% accuracy degradation.
\end{itemize}

\section{Related Work}
\subsection{Edge–Cloud Co-Inference}
Edge-cloud co-inference, or split computing, partitions a neural network between a resource-constrained edge device and a powerful cloud server to balance latency, energy, and accuracy. Early works like BottleNet and BottleNet++ demonstrated the effectiveness of inserting learned compression bottlenecks at the partition point for traditional deep neural networks (DNNs), achieving significant feature compression with minimal accuracy impact \cite{shao2020bottlenet++}. In the neuromorphic domain, approaches like EC-SNN split SNNs but offload raw, uncompressed spike features, failing to address the communication bottleneck \cite{di2024ec}. More relevantly, SpikeBottleNet and SNN-SC introduced spike-driven compression, showing massive bandwidth savings \cite{hassan2024spikebottlenet},\cite{wang2024snn}. However, these methods were often limited to low-resolution static datasets or did not address the temporal redundancy in SNNs. SNN-SC-HARQ dynamically adjusts timesteps based on channel conditions, whereas NeuCODEX adapts to input difficulty, providing a more direct optimization of computational resources \cite{wang2025spiking}.

\subsection{Dynamic Early Exit Mechanism in Neural Networks}
To reduce the high computational cost associated with deep SNNs that require many timesteps, dynamic SNNs have been proposed \cite{li2023unleashing}. These methods adapt the number of inference timesteps on a per-input basis. Techniques such as Spiking Early-Exit Neural Networks (SEENN) and dynamic confidence decoding terminate inference early once a confidence threshold is met, using either learned policies or output-based heuristics \cite{li2023seenn}. While these approaches have proven effective for SNNs in standalone settings, they have not been evaluated in scenarios involving feature encoding and subsequent reconstruction via decoder mechanisms. NeuCODEX is the first to bridge this gap, unifying temporal adaptation with spatial compression to jointly optimize the entire edge–cloud system for latency, bandwidth, and energy efficiency.

\section{Methodology}
The NeuCODEX architecture is designed to minimize latency and energy in edge-cloud SNN inference by synergistically combining two key techniques: a spike-driven compression bottleneck and a dynamic early-exit mechanism.

\subsection{NeuCODEX Architectural Overview}
\begin{figure*}[htbp]
\centering
\includegraphics[width=\linewidth]{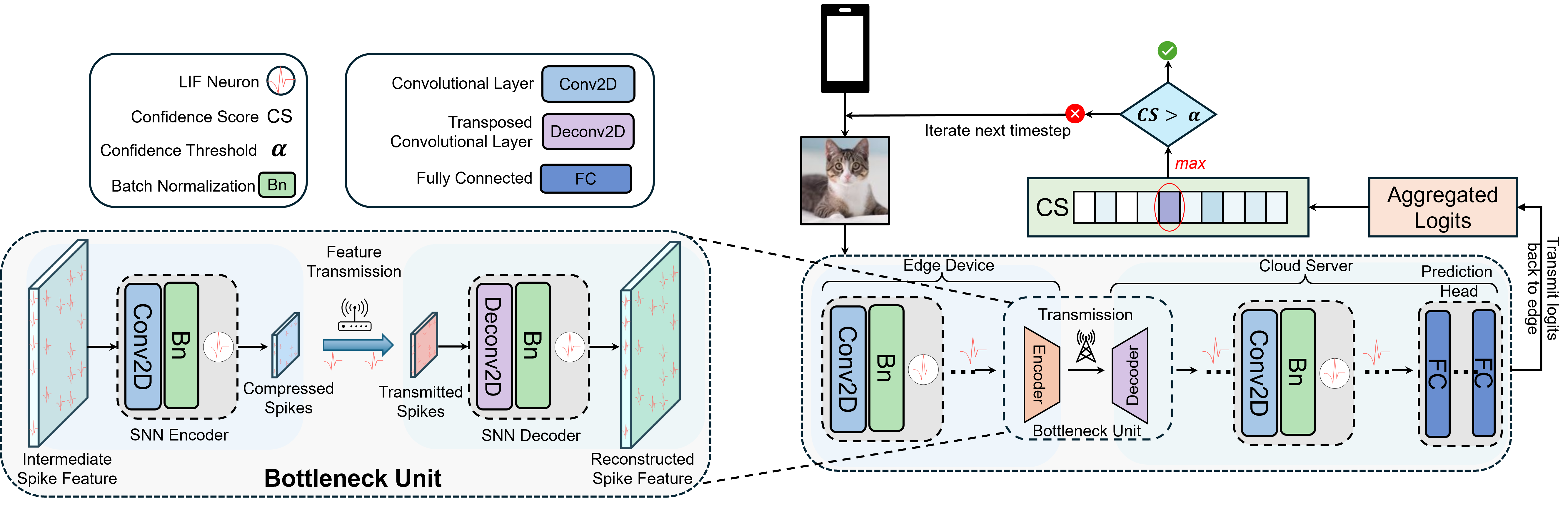}
\caption{Schematic of the NeuCODEX edge–cloud co-inference architecture. The input is processed by early SNN layers on the edge, followed by a spike-driven compression module at the partition. The compressed feature is transmitted to the cloud, where it is decoded and processed by the remaining network layers. At each timestep, a confidence-based early-exit decision at the edge can potentially terminate inference before all timesteps are used.}
\label{fig:neucodex-architecture}
\end{figure*}

As illustrated in Figure \ref{fig:neucodex-architecture}, an SNN is partitioned at a specific layer. The initial layers ($f_{edge}$) run on the edge device, while the remaining layers ($f_{cloud}$) are deployed to the cloud. At the partition point, a lightweight encoder ($E$) compresses the intermediate spike-based feature map, $X_t$, into a compact representation, $Z_t$. This compressed data is transmitted to the cloud, where a decoder ($D$) reconstructs an approximation of the feature map, $\hat{X}_t$, before it is fed into the subsequent layers. 

Crucially, after each timestep $t$, the cloud-side model computes output logits $y(t)$ and sends them back to the edge. The edge device then calculates a confidence score to determine if the prediction is sufficiently certain. If it is, inference is terminated, and the final prediction is returned. Otherwise, the process continues to the next timestep, up to a predefined maximum $T_{max}$. This dynamic, input-dependent approach avoids unnecessary computation and communication for `easy' inputs.

\subsection{Spike-Driven Bottleneck Architecture}
NeuCODEX inserts a bottleneck unit at the chosen split layer of the SNN to perform feature compression. Formally, let $X_t \in \{0,1\}^{C \times H \times W}$ denote the spike activation tensor at the partition layer for timestep $t$. Rather than transmitting the full tensor $X_t$, an encoder on the edge generates a compressed feature ${Z_t}\in \mathbb{R}^{C' \times W'\times H'}$ (where $C'\ll C, W'\ll W, H'\ll H$). 

The encoder function, $f_{enc}$, is composed of a 2D convolution ($Conv$), batch normalization ($BN$), and a Leaky-Integrate-and-Fire (LIF) neuron ($\mathcal{SN}$), such that:

\begin{align}
\mathbf{Z}_t = f_{enc}(\mathbf{X}_t) = \mathcal{SN} \left( BN \left( Conv(\mathbf{X}_t, W_{enc}) \right) \right)
\end{align}
where $W_{enc}$ represents the learned weights of the encoder's convolutional layer. Although SpikeBottleNet originally adopts time-dependent batch normalization  (${\rm{tdBN}}$), prior studies have shown that conventional  ${\rm{BN}}$ is sufficient for training deep SNNs. Therefore, we replace ${\rm{tdBN}}$ with the standard ${\rm{BN}}$.

The compressed features $Z_t$ are transmitted as a small binary packet over the network. On the cloud side, a corresponding decoder reconstructs an approximation $\hat{X}_t$ of the original spike tensor using a decoder function, $f_{dec}$:
\begin{align}
\hat{\mathbf{X}}_t = f_{dec}(\mathbf{Z}_t) = \mathcal{S} \left( BN \left( ConvTranspose(\mathbf{Z}_t, W_{dec}) \right) \right)
\end{align}

where $ConvTranspose$ is a 2D transposed convolution and $W_{dec}$ are the learned weights of the decoder. The encoder–decoder pair constitutes the bottleneck unit and is illustrated in the Figure \ref{fig:neucodex-architecture}.

\textbf{LIF neuron in NeuCODEX}. The membrane potential dynamics of an LIF neuron (${\mathcal{SN}}$) at time step $t$ are governed by the following equations:

\begin{equation}
H[t] = V[t - 1] + \frac{1}{\tau} \left( X[t] - (V[t - 1] - V_{\text{reset}}) \right),
\end{equation}
\begin{equation}
S[t] = \Theta(H[t] - V_{\text{th}}),
\end{equation}
\begin{equation}
V[t] = H[t](1 - S[t]) + V_{\text{reset}} S[t],
\end{equation}

where $\tau$ is the membrane time constant and $X[t]$ denotes the input current at time step $t$. A spike is generated if the membrane potential $H[t]$ exceeds the threshold $V_{\text{th}}$, which is determined using the Heaviside step function $\Theta(v)$ that returns 1 when $v \geq 0$ and 0 otherwise. After a spike, the membrane potential resets to $V_{\text{reset}}$; otherwise, it remains at $H[t]$. We used the SpikingJelly's $\tau = 2.0,\;V_{th} = 1.0,\;V_{\text{reset}} = 0.0$ default parameters \footnote{https://spikingjelly.readthedocs.io/zh-cn/latest/}.

\subsection{Dynamic Early-Exit Strategy}
NeuCODEX implements a dynamic timestep control mechanism to exploit temporal redundancy and minimize unnecessary computation and communication in edge–cloud SNN deployments.

Let the SNN operate for at most \( T_{\text{max}} \) timesteps. At each timestep 
\( t \), the edge processes the input to produce spike features \( Z_t \), which are compressed and transmitted to the cloud. The cloud decodes these and computes class logits \( \mathbf{y}^{(t)} \in \mathbb{R}^K \), which are sent back to the edge, where \( K \) is the number of classes. We compute the probability vector \( \mathbf{p}^{(t)} = \text{softmax}(\mathbf{y}^{(t)}) \), and define the confidence score as:

\begin{equation}
\text{CS}^{(t)} = \max_{i \in \{1, \ldots, K\}} p_i^{(t)}
\label{eq:confidence_score}
\end{equation}

Inference terminates at the earliest timestep \( t_{\text{exit}} \) such that \( \text{CS}^{(t)} \geq \alpha \), where \( \alpha \in (0, 1) \) is a predefined confidence threshold:

\begin{equation}
t_{\text{exit}} = \min \left\{ t \in \{1, \ldots, T_{\text{max}} \} \mid \text{CS}^{(t)} \geq \alpha \right\}
\label{eq:exit_time}
\end{equation}

If no such \( t \) exists, the model defaults to using all \( T_{\text{max}} \) timesteps. The final prediction is \( \hat{k} = \arg\max_i p_i^{(t_{\text{exit}})} \).

This strategy allows NeuCODEX to skip redundant timesteps for easier inputs, reducing both edge compute and uplink communication. The threshold \( \alpha \) governs the trade-off between latency and accuracy, and can be tuned based on dataset-level tolerances. In our experiments, values of \( \alpha \in [0.90, 0.95] \) achieve an effective latency–accuracy balance across datasets.

\renewcommand{\arraystretch}{1.3} 
\setlength{\tabcolsep}{10pt}      

\begin{table*}[htbp]
\caption{Accuracy-T comparison of ResNet-18 and VGG-16 on various datasets. Each cell shows the accuracy (\%) followed by the timestep (T) and the change in accuracy ($\Delta$) relative to the fixed-timestep no-bottleneck baseline (F–B), formatted as (T, $\Delta$).}
\centering
\begin{tabular}{|l|c|c|c|c|c|}
\hline
\textbf{Models} & \textbf{Config.} & \textbf{CIFAR10 (T, $\Delta$)} & \textbf{CIFAR10-DVS (T, $\Delta$)} & \textbf{Caltech (T, $\Delta$)} & \textbf{NCaltech (T, $\Delta$)} \\
\hline
\multirow{4}{*}{ResNet-18}
  & F-B & 94.15 (2, 0) & 65.8 (2, 0) & 97.92 (2, 0) & 94.25 (2, 0) \\
  & D-B & 93.62 (1.18, -0.53) & 63.0 (1.28, -2.8) & 97.70 (1.68, -0.22) & 93.36 (1.05, -0.89) \\
  & F+B & 91.46 (2, -2.69) & 64.55 (2, -1.25) & 96.13 (2, -1.79) & 92.63 (2, -1.62) \\
  & D+B & 92.63 (1.16, -1.52) & \textbf{67.4 (1.18, +1.75)} & 97.62 (1.3, -0.3) & 93.95 (1.04, -0.3) \\
\hline
\multirow{4}{*}{VGG-16}
  & F-B & 91.35 (2, 0) & 63.95 (2, 0) & 97.92 (2, 0) & 91.45 (2, 0) \\
  & D-B & \textbf{91.37 (1.08, +0.02)} & 63.2 (1.76, -0.75) & 97.32 (1.22, -0.6) & 89.23 (1.03, -2.22) \\
  & F+B & 88.51 (2, -2.84) & 62.75 (2, -1.2) & 97.17 (2, -0.75) & \textbf{92.03 (2, +0.58)} \\
  & D+B & 90.09 (1.09, -1.26) & 62.8 (1.72, -1.15) & 97.17 (1.71, -0.75) & 89.23 (1.02, -2.22) \\
\hline
\end{tabular}
\label{table:spike-network-accuracy}
\end{table*}

\subsection{Training Strategy}
We adopt spiking ResNet-18 and VGG-16 backbones, constructed by LIF neurons after each convolutional and fully connected layer. The base SNNs are trained with surrogate gradient descent over $T = 2$ timesteps to minimize the cross-entropy loss $\mathcal{L}_{\text{CE}}$. After convergence of the base SNNs, we insert a compression bottleneck at a candidate splitting point, yielding the modified network. The encoder and decoder of the bottleneck are initialized randomly. Importantly, the bottleneck compression is lossy – i.e., $\hat{X}_t$ is not exactly $X_t$. To compensate for the information loss from compression, the entire SNN is fine-tuned using the cross-entropy loss ($\mathcal{L}_{\text{CE}}$) after the bottleneck is inserted.

\textbf{Partition Variants.}
We evaluate multiple split points $\{l_s^{(i)}\} = \{\text{SP1}, \text{SP3}, \text{SP5}, \text{SP7}, (\text{SP8 for VGG})\}$, corresponding to increasing feature sizes $C_i$ and spatial dimensions $H_i \times W_i$, where higher $i$ implies greater transmission costs.

\textbf{Early-Exit Integration.}
Finally, NeuCODEX integrates the dynamic early-exit mechanism as described in Sec. III.C, introducing an adaptive exit timestep $t_{\text{exit}} \leq T$ based on confidence threshold $\alpha \in (0, 1)$. This further reduces average compute and transmission cost per sample.

Overall, NeuCODEX is designed to minimize all three components of inference latency: edge compute, data transmission, and cloud compute, resulting in significant end-to-end speedups. We note that with a higher $T_{\max}$ (e.g. 8 timesteps), the benefits of early exit would be even more pronounced on average.

\section{Experimental Results}
\subsection{Datasets, Models, and Setup}
For static images, we use CIFAR-10, a dataset with 10 classes of $32\times32$ pixel RGB images. For neuromorphic data, CIFAR10-DVS, which is a Dynamic Vision Sensor (DVS) event-based version of CIFAR-10, containing asynchronous spike event streams capturing the same image classes, is used. Additionally, to test on a high-resolution vision task, we include a 10-class subset of Caltech101, and its neuromorphic counterpart NCaltech101.

\textbf{Spike encoding}. We employ rate coding to simulate spiking input from static images. Each pixel is converted into a short spike train spanning two timesteps, with higher pixel intensities producing higher firing rates over those timesteps. The neuromorphic datasets are already in spike event form. In all cases, the SNN processes 2 timesteps in the baseline fixed-timestep scenario ($T_{max}$ = 2).

\textbf{Hardware and measurements}. Our hardware testbed consists of an NVIDIA Jetson Orin Nano\footnote{https://www.nvidia.com/en-us/autonomous-machines/embedded-systems/jetson-orin/nano-super-developer-kit/} as the edge device and an NVIDIA A100-SXM4\footnote{https://www.nvidia.com/en-us/data-center/a100/} GPU server as the cloud. They communicate over a standard Wi-Fi network with \textasciitilde 18.9 Mbps measured throughput.

Four configurations are compared: \textbf{(F–B)} Fixed timesteps, no bottleneck (baseline SNN with 2 timesteps, transmitting full spike tensors); \textbf{(D–B)} Dynamic timesteps, no bottleneck (early-exit enabled but still sending full uncompressed tensors); \textbf{(F+B)} Fixed timesteps with bottleneck compression; and \textbf{(D+B)} NeuCODEX, dynamic timesteps with bottleneck, our full method. This breakdown helps us to isolate the effects of compression and dynamic inference, both separately and in combination.

\begin{figure}[ht]
\centering
\includegraphics[width=\linewidth, keepaspectratio]{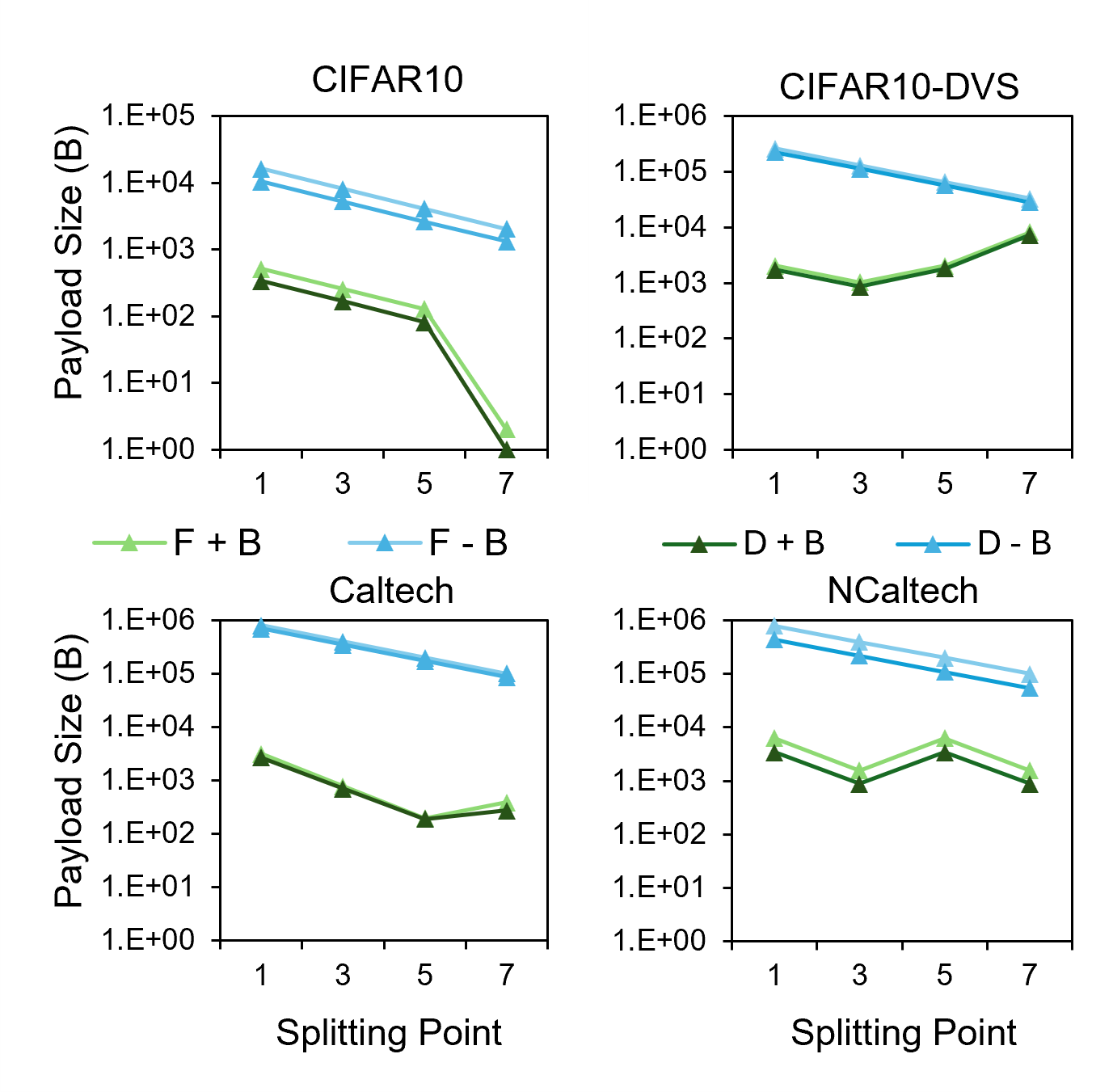}
\caption{Payload size (in bytes) of the data transmitted per inference by NeuCODEX’s bottleneck encoder at various edge–cloud split points, across different datasets. Lower values indicate greater data reduction. F = Fixed Timestep, D = Dynamic Timestep, +B = With Bottleneck, –B = Without Bottleneck.}
\label{fig:compression-capability}
\end{figure}

\subsection{Accuracy and Early-Exit Behavior}
As shown in Table \ref{table:spike-network-accuracy}, NeuCODEX (D+B) maintains high accuracy while significantly reducing the average number of timesteps $(T_{avg})$. Across all datasets and models, the accuracy drop is minimal, typically between 1-2\% compared to the 2-timestep baseline (F-B), and in some cases, such as CIFAR10-DVS, accuracy even improves slightly. This is achieved with a confidence threshold $\alpha$=0.9, where 50-70\% of samples exit after the first timestep. This demonstrates that the early-exit mechanism effectively identifies `easy' inputs without sacrificing performance on more difficult ones.

\subsection{Bandwidth and Communication Savings}
The spike compression of NeuCODEX results in significant communication overhead reductions, as highlighted in Figure \ref{fig:compression-capability}. In the baseline (F–B) scenario, the edge must send a raw spike tensor of shape $(T, C, H, W)$. For SNNs these are binary spikes (so one bit per neuron), but the sheer number of neurons can make even a sparse spike map large in the worst case. With our bottleneck (B), the edge instead sends a much smaller compressed code per timestep. Moreover, thanks to dynamic early exit (D), on average fewer timesteps are transmitted per input.

For instance, on CIFAR-10 with ResNet-18 at the deepest split point 7 (SP7), the data payload per sample is compressed from 16,384 bits to just 8 bits, a 2048x reduction. Even for the higher-resolution Caltech dataset, a 1056x reduction is achieved. This makes inference feasible even over low-bandwidth networks.

\subsection{End-to-End Latency Results}
\begin{figure}[ht]
\centering
\includegraphics[width=\linewidth, keepaspectratio]{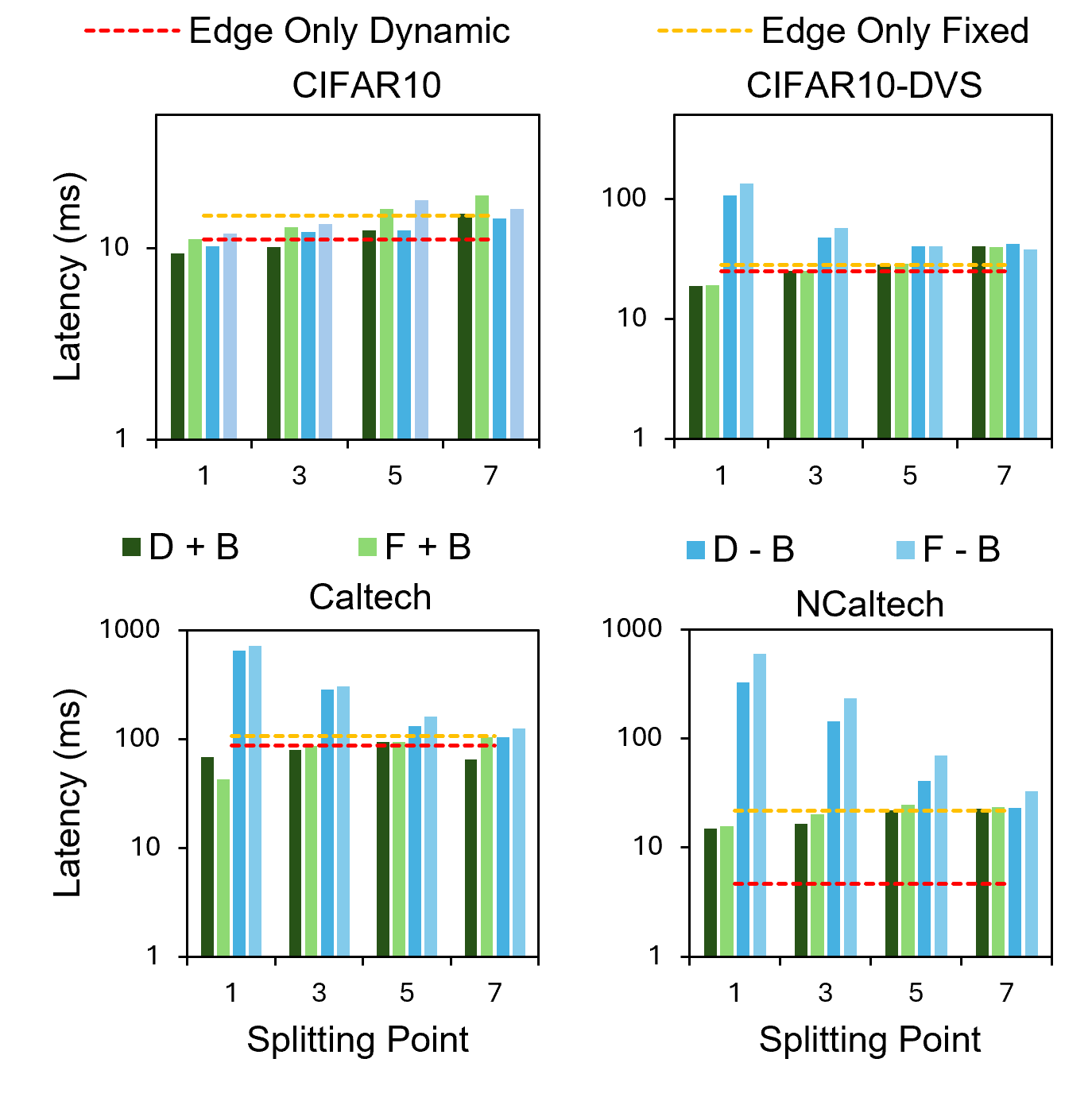}
\caption{End-to-end inference latency for ResNet-18 (including edge computation, transmission, and cloud processing) for baseline vs. NeuCODEX configurations across multiple datasets. NeuCODEX consistently achieves lower total latency compared to the baseline. F = Fixed Timestep, D = Dynamic Timestep, +B = With Bottleneck, –B = Without Bottleneck}
\label{fig:resnet_latency_measurement}
\end{figure}

Reducing communication and computation yields lower overall inference latency, which is critical in edge scenarios. We measured the end-to-end latency per sample, including edge processing, network transmission, and cloud processing. As presented in Figure \ref{fig:resnet_latency_measurement}, NeuCODEX significantly outperforms the baseline's edge-only processing, especially when the network is split early (where the uncompressed data would have been largest). For example, on the N-Caltech dataset with an early split point (SP1), NeuCODEX reduces latency by 31.8\% compared to a full edge-only SNN. Compared to naive uncompressed transmission, NeuCODEX was over 95\% faster (597 ms → 15 ms), demonstrating the infeasibility of uncompressed SNN splitting. The speedups come from sending far less data (e.g., 590 ms of transmission cut down to \textasciitilde 7 ms in one case) and from avoiding extra timesteps on the edge due to early exits.

\subsection{Energy Consumption at Edge} 
Energy efficiency is a major advantage of SNNs, and NeuCODEX furthers this by reducing the amount of spiking activity and computation needed on the edge device. We theoretically estimate the edge energy per inference using a model of the Intel Loihi neuromorphic chip’s power: \textasciitilde23 pJ per synaptic operation (SynOp) \cite{davies2018loihi}. 

\begin{figure}[ht]
\centering
\includegraphics[width=\linewidth, keepaspectratio]{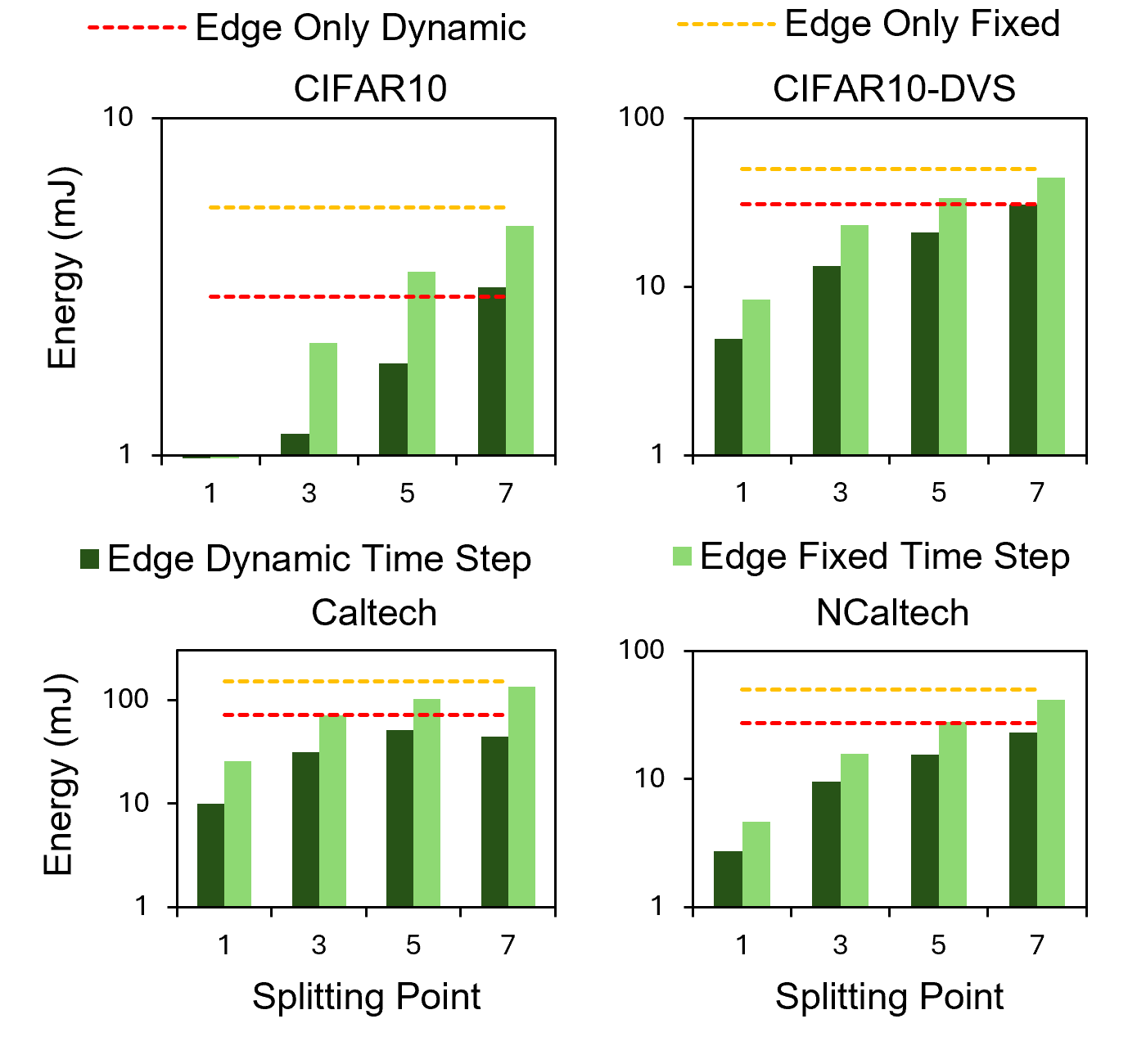}
\caption{Estimated edge-device energy consumption per inference at different network partition points for a ResNet-18 SNN across various datasets. NeuCODEX greatly lowers the required energy on the edge by offloading computation after early layers and by reducing the number of timesteps processed locally.}
\label{fig:resnet_energy_consumption}
\end{figure}

The results for ResNet-18 are shown in Figure \ref{fig:resnet_energy_consumption}. By offloading the bulk of the computation and reducing the active processing time via early exits, NeuCODEX reduces the estimated computational energy draw on the edge device by 90-94\% across all tested scenarios. For example, on CIFAR-10, energy consumption dropped from 5.45 mJ to 0.43 mJ, and on N-Caltech, from 49.90 mJ to 2.75 mJ. This makes it highly suitable for battery-powered IoT devices.

\section{Ablation Study}
We conducted an ablation study to examine the impact of the bottleneck and the dynamic timestep mechanism under different settings. This study was performed on the CIFAR-10 dataset using the ResNet-18 (SP3) backbone on edge CPU, with a deeper maximum timesteps budget to illustrate the trends.

\begin{figure}[ht]
\centering
\includegraphics[width=2.5in, keepaspectratio]{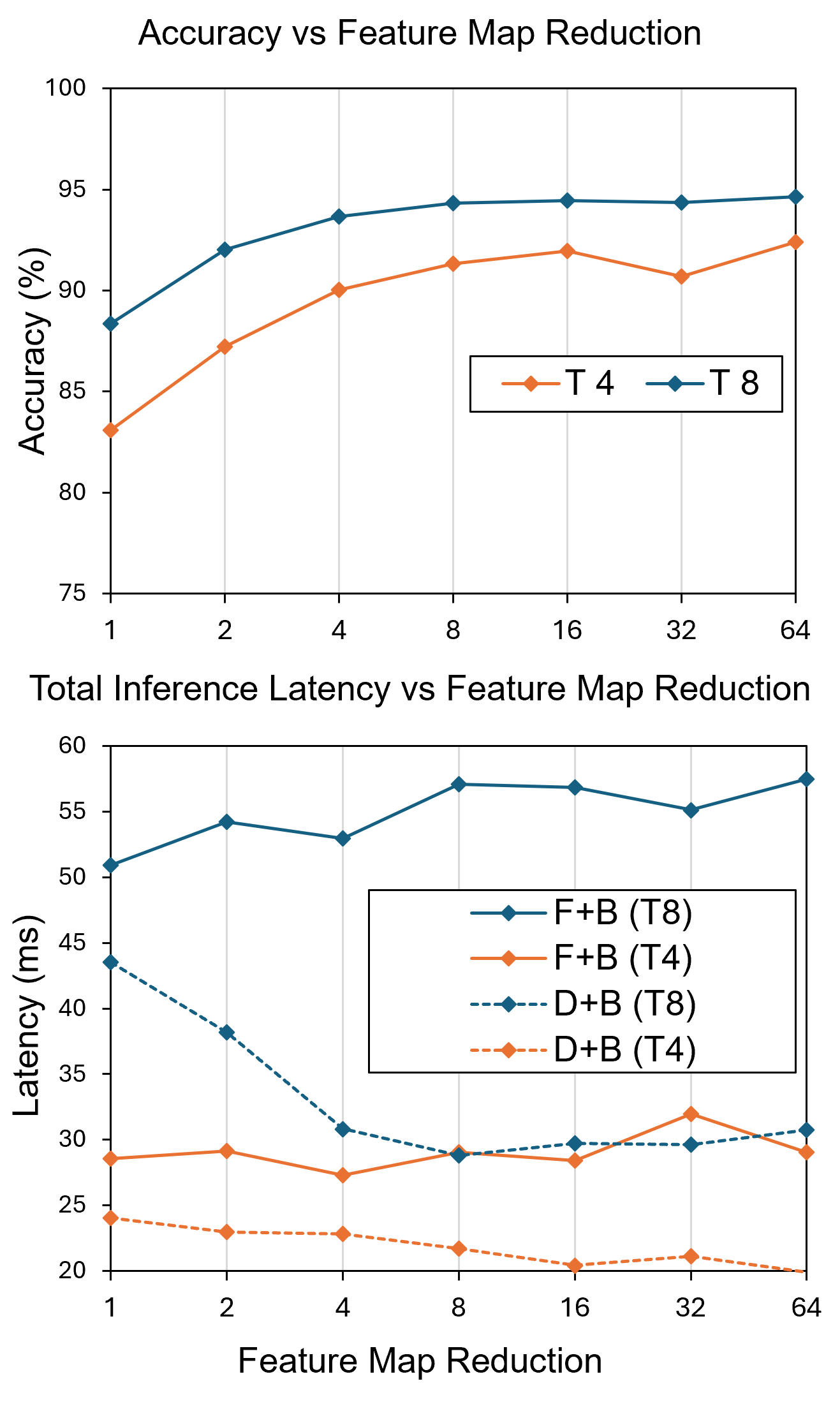}
\caption{Ablation study on CIFAR-10 (ResNet-18, SP3) showing how increasing the maximum timestep budget ($T_{\max}=4$ vs $T_{\max}=8$) affects accuracy and latency for NeuCODEX under different compression settings (varying bottleneck channel count). F = Fixed Timestep, D = Dynamic Timestep, +B = With Bottleneck}
\label{fig:resnet_ablation}
\end{figure}

Figure \ref{fig:resnet_ablation} summarizes the findings. We observe that allowing up to 8 timesteps (vs 4) gave the model more temporal budget for hard inputs, raising accuracy (e.g. 85.1\% → 88.3\%) while reducing average latency (43.5 → 28.6 ms) due to more frequent early exits. The ablation also showed that too aggressive compression (very few bottleneck channels) hurts accuracy more than it helps latency, whereas a slightly larger bottleneck yields better accuracy with only a minor latency penalty.

\section{Conclusion}
This work presents NeuCODEX, a unified neuromorphic co-inference architecture designed to address the bandwidth, latency, and energy bottlenecks in edge-to-cloud SNN deployments. NeuCODEX employs a learned spike compression module at the edge-cloud interface and a dynamic confidence-adaptive early exit mechanism to avoid redundant timesteps and reduce data transmission. Experimental results using deep spiking models across multiple image recognition datasets reveal that NeuCODEX achieves substantial improvements in communication efficiency, edge-side energy savings, and total inference speed while preserving model accuracy. These findings underscore the importance of jointly optimizing spatial and temporal dimensions in delivering lightweight, high-performance SNN inference on resource-limited platforms like IoT devices and autonomous agents. Potential extensions of this work include applying NeuCODEX to other spiking network architectures (such as Vision Transformer) and to tasks like object detection, image segmentation, or image super-resolution, as well as exploring hardware-level implementations on neuromorphic chips.

\section{Acknowledgment}
This publication has emanated from research conducted with the financial support of Taighde Éireann – Research Ireland under Grant number 18/CRT/6222 and 21/FFP-A/9174. The authors would like to acknowledge Huawei Ireland Research Centre for providing the computing resources that supported this research.

\clearpage
\setcounter{page}{1}

\twocolumn[
\centering
\Large
\textbf{NeuCODEX: Edge–Cloud Co-Inference with Spike-Driven Compression and Dynamic Early-Exit}\\
\vspace{0.5em}Supplementary Material \\
\vspace{1.0em}
]

\section{ResNet-18 Model Analysis}
This section provides a detailed analysis of the ResNet-18 results presented in the main paper, elaborating on the accuracy, bandwidth, latency, and energy performance.
\subsection{Accuracy and Early-Exit Behavior (ResNet-18)}
Across all datasets and models, NeuCODEX maintains accuracy very close to the original 2-timestep SNN (F-B), despite the lossy compression and using fewer timesteps. For example, on N-Caltech, the baseline ResNet-18 (fixed 2 timesteps, no compression) achieved 94.25\%. With NeuCODEX (at SP1), it attained 93.95\% – a drop of merely 0.3\%. A similar tiny drop was seen on high-res Caltech (baseline 97.92\% vs 97.62\% with NeuCODEX). On CIFAR-10, NeuCODEX (ResNet-18, SP1) reached 92.63\% vs 93.62\% for a baseline that used dynamic inference on the edge without compression (D-B). This \textasciitilde 1-2\% difference is acceptable given the huge gains in efficiency (as detailed below). Interestingly, on the CIFAR10-DVS neuromorphic dataset, NeuCODEX slightly improved accuracy compared to some baselines: e.g. 67.4\% at SP3 for ResNet-18 (D+B) vs fixed baselines of 64–66\%.

Overall, the accuracy impact of NeuCODEX is negligible – typically within 1\% of the baseline – confirming that our compression bottleneck preserved essential information and that the confidence threshold was set conservatively  (we used $\alpha$ between 0.90 and 0.95) enough to avoid sacrificing accuracy for speed. The dynamic early exit does not degrade accuracy because if a sample is not sufficiently confident, it simply uses the full timesteps (thus matching baseline performance). In fact, in our experiments, roughly 50–70\% of samples (varies by dataset) exited after 1 timestep when $\alpha=0.9$, while the rest took 2 steps; this adaptive process did not significantly misclassify any samples that the full 2-step run would have gotten correct.

\subsection{Bandwidth and Communication Savings (ResNet-18)}
The spike compression of NeuCODEX results in significant communication overhead reductions, as highlighted in Figure \ref{fig:compression-capability}. In the baseline (F–B) scenario, the edge must send a raw spike tensor of shape $(T, C, H, W)$. For SNNs these are binary spikes (so one bit per neuron), but the sheer number of neurons can make even a sparse spike map large in the worst case. With our bottleneck (B), the edge instead sends a much smaller compressed code per timestep. Moreover, thanks to dynamic early exit (D), on average fewer timesteps are transmitted per input.

As a concrete example: on CIFAR-10 with ResNet-18 at the deepest split (SP7), the raw feature tensor would be 16,384 bits per sample (2 timesteps of a $512\times4\times4$ spike feature). NeuCODEX’s encoder compressed this to just 8 bits (1 byte) per sample – a $2048\times$ reduction. If an early exit triggers at $t=1$, the second timestep’s code is never sent. In practice, we observed an average of 1.18 encoded tensors sent instead of 2 on CIFAR-10 (ResNet-18, D+B), since \textasciitilde 82\% of inputs exited after the first step. 

Similar trends were found on other tasks: for CIFAR10-DVS, a raw stream of \textasciitilde 131k bits was compressed to 854 bits (\textasciitilde 154×) at SP3; for high-res Caltech, \textasciitilde 200k bits reduced to 190 bits (\textasciitilde 1056×) at SP5. Even at more moderate splits like SP3 on NCaltech, we achieved \textasciitilde 458× compression. In all cases, NeuCODEX (D+B) transmitted orders-of-magnitude fewer bits than the uncompressed baseline, often 3–4 orders less. This drastic reduction directly translates to lower network latency.

\subsection{End-to-End Latency Results (ResNet-18)}
Reducing communication and computation yields lower overall inference latency, which is critical in edge scenarios. We measured the end-to-end latency per sample, including edge processing, network transmission, and cloud processing. As presented in Figure \ref{fig:resnet_latency_measurement}, NeuCODEX significantly outperforms the baselines edge-only processing, especially when the network is split early (where the uncompressed data would have been largest). For example, on NCaltech (which has large $224\times224$ spike frames), running the full SNN on the Jetson Orin (edge-only inference) took 21.79 ms. When we split at SP1 and applied NeuCODEX, the total latency (edge + WiFi + cloud) dropped to 14.85 ms – a 31.8\% improvement over edge-only, even though we now incur transmission and cloud processing. This is significant, and is due to the fact that the edge-only case had to process two full timesteps of the entire network, whereas NeuCODEX’s split offloads the majority of layers to the fast cloud GPU and also often stops after 1 timestep. Compared to the extremely slow scenario of sending uncompressed features at SP1 (which took an enormous 597.29 ms total) NeuCODEX’s 14.85 ms is a 97\% reduction in latency. Of course, that 597.29 ms is mostly transmission time for huge tensors – a regime one would typically avoid – but it highlights how infeasible uncompressed split-SNN would be on high-res data. NeuCODEX makes such splitting not only feasible but actually faster than local processing by more than $2\times$. 

On Caltech (static images, high-res), NeuCODEX at SP1 achieved 68.04 ms total latency vs 107.19 ms edge-only (a 36.5\% speedup). On CIFAR10-DVS, the gains were smaller but still notable: edge-only was 28.05 ms, NeuCODEX at SP3 was 25.06 ms (about 11\% faster). Finally, even in a scenario where the edge uses a CPU (simulating a very low-end device), NeuCODEX provided improvements: for example, on CIFAR-10, edge-only CPU latency 14.86 ms vs NeuCODEX 9.39 ms (37\% speed-up).

\begin{figure}[ht]
\centering
\includegraphics[width=\linewidth, keepaspectratio]{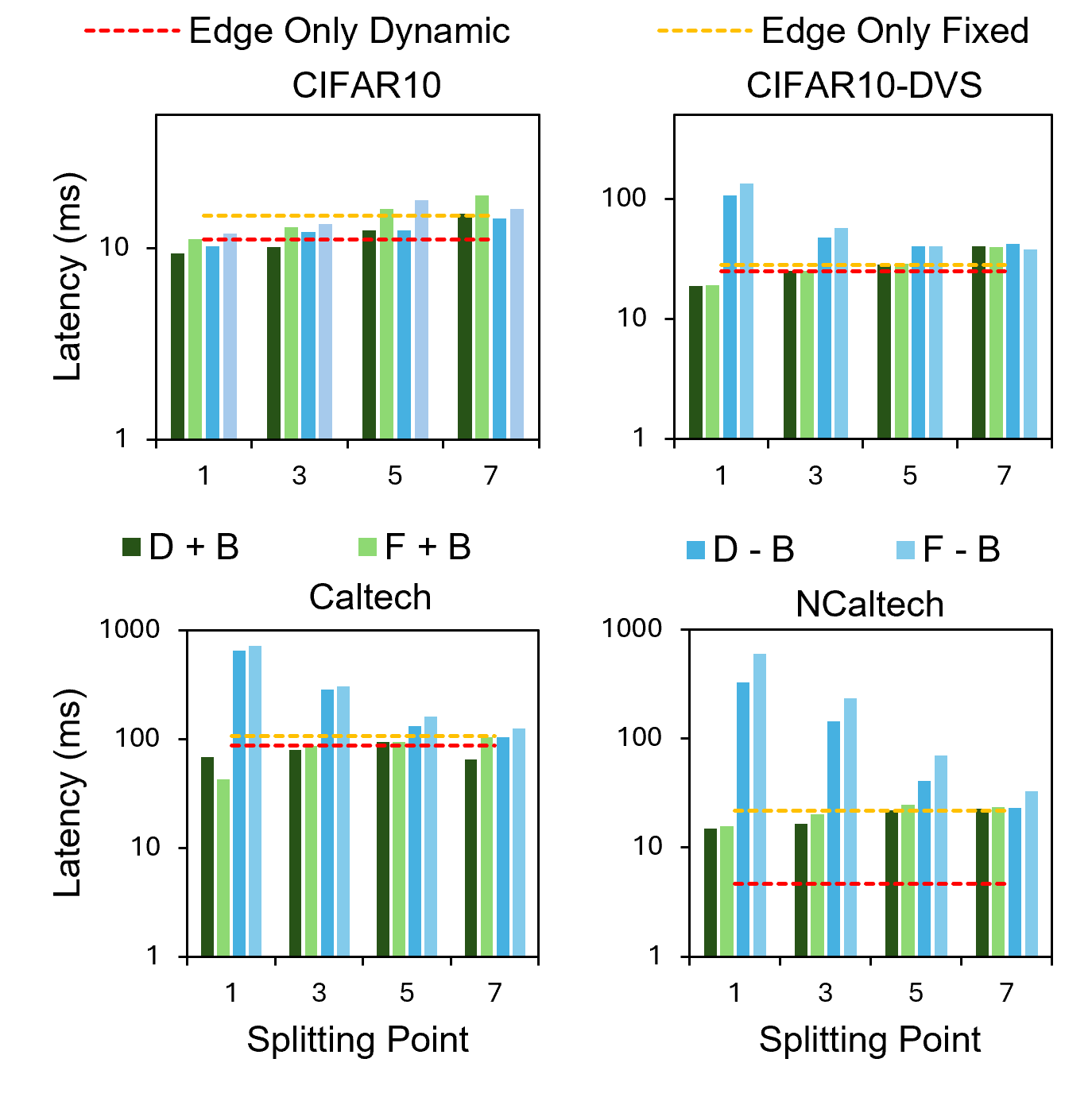}
\caption{End-to-end inference latency for ResNet-18 (including edge computation, transmission, and cloud processing) for baseline vs. NeuCODEX configurations across multiple datasets. NeuCODEX consistently achieves lower total latency compared to the baseline. F = Fixed Timestep, D = Dynamic Timestep, +B = With Bottleneck, –B = Without Bottleneck}
\label{fig:latency-measurement}
\end{figure}

It is instructive to break down where these latency gains come from. First, transmission time is cut drastically by compression: e.g. at NCaltech SP1, sending the uncompressed feature took \textasciitilde 590 ms, whereas NeuCODEX’s compressed feature took only \textasciitilde 7.26 ms – this \textasciitilde 583 ms saved is huge. Second, cloud computation on the small decoded features is very fast (the cloud runs layers on a powerful GPU, and it’s marginal compared to transmission in most cases). Third, the edge computation time per timestep in NeuCODEX is a bit lower than in the edge-only case because the edge stops at the split (does not execute the remaining layers). However, this per-timestep saving can be offset if we end up doing more timesteps. That is where the dynamic early exit helps: by cutting the average number of timesteps, we further reduce the edge compute load. For instance, in the N-Caltech case above, NeuCODEX only averaged \textasciitilde 1.05 timesteps (due to many 1-step exits), effectively doing just over half the work of a full 2-timestep run on the edge. Thus, edge compute time also went down proportionally. In summary, NeuCODEX minimizes all three components of latency: edge compute, data transmission, and cloud compute, yielding a significant net speedup.

\subsection{Edge Energy Consumption (ResNet-18)}
Energy efficiency is a major advantage of SNNs, and NeuCODEX furthers this by reducing the amount of spiking activity and computation needed on the edge device. We theoretically estimate the edge energy per inference using a model of the Intel Loihi neuromorphic chip’s power: \textasciitilde23 pJ per synaptic operation (SynOp) \cite{davies2018loihi}. We calculate the number of SynOps for each layer by multiplying its FLOPs by the layer’s firing rate and number of timesteps (since not all neurons fire) \cite{hassan2024spikebottlenet} \cite{hu2021spiking}. This gives us a proxy for how much energy the edge would spend on the SNN portion it runs. Note that cloud energy is not a concern in our setup (the server has ample power and we assume it can handle the extra load).

We compare the baseline edge-only full SNN vs NeuCODEX at various splits. The results for ResNet-18 are shown in Figure \ref{fig:resnet_energy_consumption}. NeuCODEX reduced the edge energy by around 90–94\% across all datasets. For CIFAR-10, the estimated energy dropped from 5.45 mJ to 0.43 mJ (\textasciitilde 92\% less); for CIFAR10-DVS, from 49.83 mJ to 4.90 mJ (\textasciitilde 90\% less); for Caltech, from 152.76 mJ to 9.93 mJ (\textasciitilde 93\% less); and for NCaltech, from 49.90 mJ to 2.75 mJ (\textasciitilde 94\% less).

\section{VGG-16 Model Analysis}
\subsection{Accuracy and Early-Exit Behavior (VGG-16)}
NeuCODEX’s early-exit strategy maintains VGG-16’s classification accuracy close to the full two-timestep baseline across all datasets. For example, on the static CIFAR10 benchmark, the baseline two-timestep accuracy is 91.4\%, whereas NeuCODEX achieves about 88.5–90.1\% (within 2.5\% of baseline). On Caltech, baseline 97.47\% (dynamic) versus 97.77\% (fixed) ($<$1\% drop). On NCaltech, baseline 97.47\% versus NeuCODEX \textasciitilde 95.24–97.17\% (\textasciitilde 2.2\% drop), and on CIFAR10-DVS, baseline 63.20\% versus NeuCODEX ~61.2–61.8\% ($<$2\% drop).

The early-exit analysis confirms that most inferences terminate at the first timestep: \textasciitilde 66–97\% of samples exit at the first early-exit in static benchmarks (with \textasciitilde 18–26\% at the second). In the DVS data, 0\% exit at the first timestep and the rest at the second (due to sparse initial spikes). The mean timesteps per inference under dynamic mode are \textasciitilde 1.08–1.13 for static datasets and \textasciitilde 1.02 for N-Caltech (vs 2 baseline), rising to \textasciitilde 1.69 for CIFAR10-DVS. Thus, NeuCODEX typically processes far fewer timesteps on average without significantly affecting accuracy.

\subsection{Bandwidth and Communication Savings (VGG-16)}
\begin{figure}[ht]
\centering
\includegraphics[width=\linewidth, keepaspectratio]{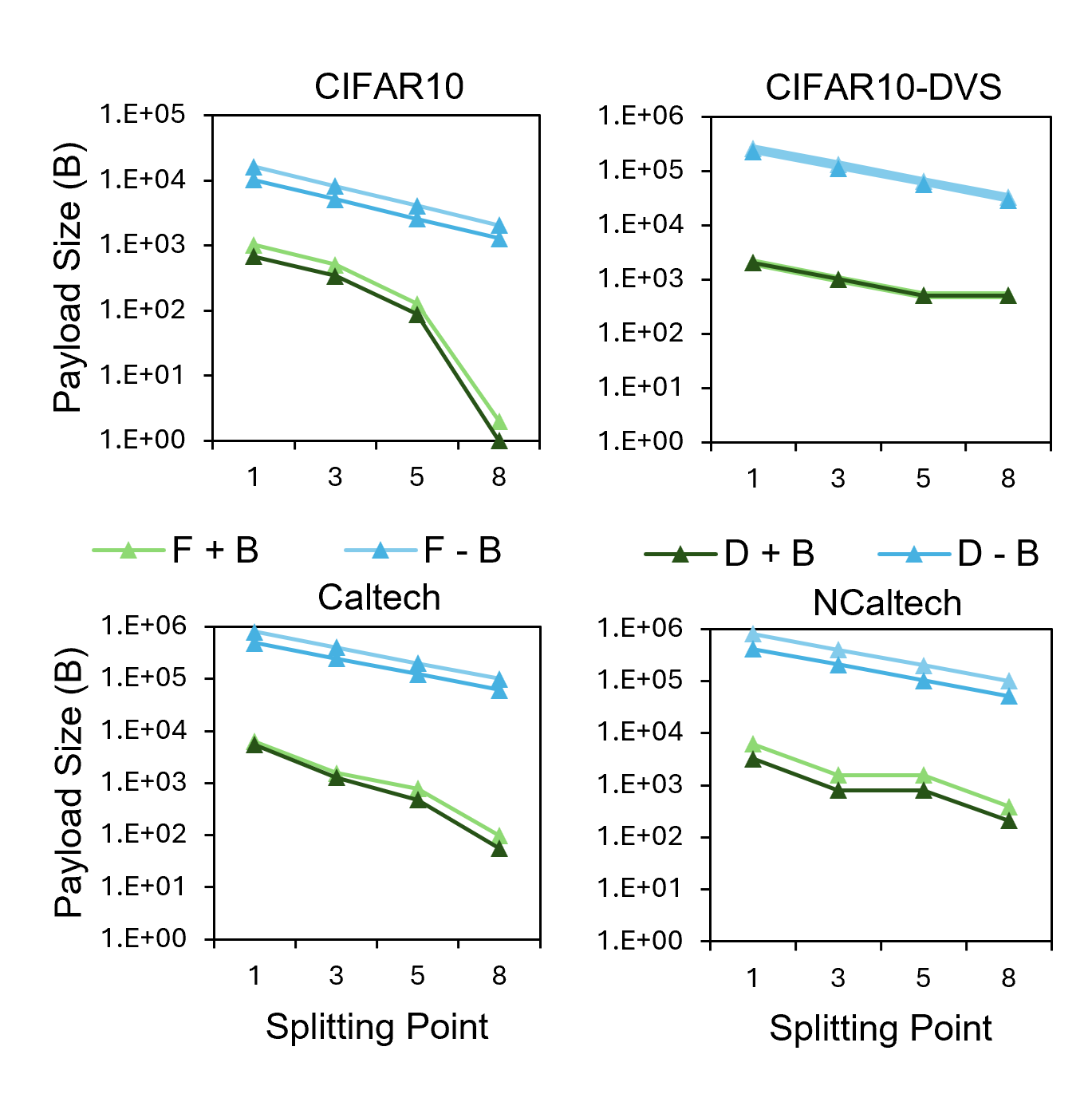}
\caption{Payload size (in bytes) of the data transmitted per inference by NeuCODEX’s bottleneck encoder at various edge–cloud split points, across different datasets. (Lower values indicate greater data reduction). F = Fixed Timestep, D = Dynamic Timestep, +B = With Bottleneck, –B = Without Bottleneck}
\label{fig:vgg_compression_capability}
\end{figure}

Figure \ref{fig:vgg_compression_capability} illustrates the results. On CIFAR-10 with VGG-16 at the deepest split (SP8), the raw spike tensor (\textasciitilde 2048 bytes for 2 timesteps) is compressed to just 2 bytes (\textasciitilde 1024× reduction) by NeuCODEX’s encoder. If an early exit triggers at $t=1$, the second timestep’s data need not be sent. Similar trends hold across all datasets: on CIFAR10-DVS, \textasciitilde 131k raw bytes are reduced to \textasciitilde 1.02k bytes (128× at SP3); on high-resolution Caltech, \textasciitilde 100k bytes are reduced to \textasciitilde 98 bytes (\textasciitilde 1024× at SP8); and on N-Caltech, \textasciitilde 400k bytes are reduced to \textasciitilde 1.57k bytes (256× at SP3). Thus, NeuCODEX transmits orders-of-magnitude fewer bits per inference than the uncompressed baseline, reflecting dramatic bandwidth savings.

\subsection{End-to-End Latency Results (VGG-16)}
\begin{figure}[ht]
\centering
\includegraphics[width=\linewidth, keepaspectratio]{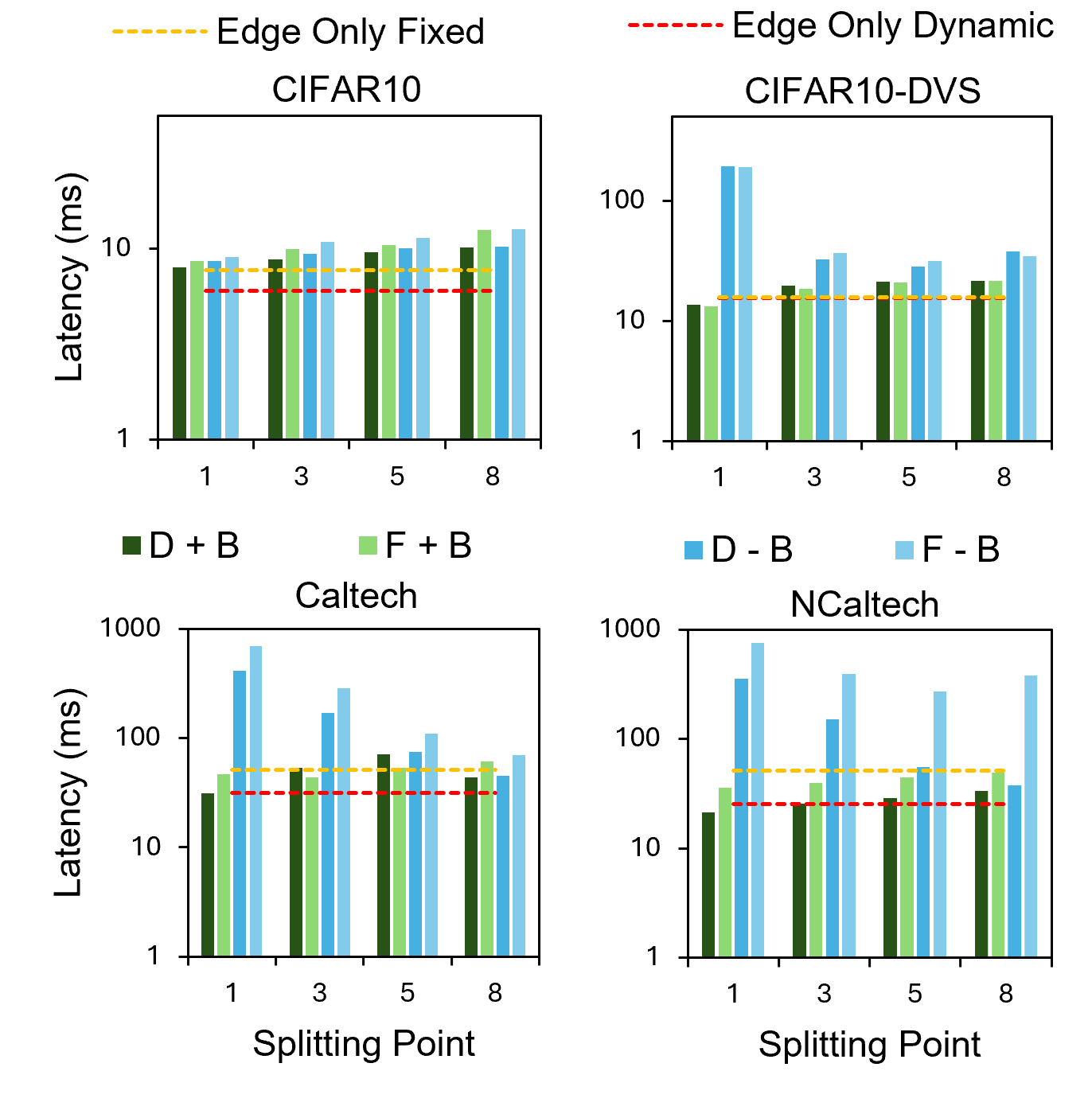}
\caption{End-to-end inference latency for VGG-16 (including edge computation, transmission, and cloud processing) for baseline vs. NeuCODEX configurations across multiple datasets. NeuCODEX consistently achieves lower total latency compared to the baseline. F = Fixed Timestep, D = Dynamic Timestep, +B = With Bottleneck, –B = Without Bottleneck}
\label{fig:vgg_latency_measurement}
\end{figure}

As shown in Figure \ref{fig:vgg_latency_measurement}, across all VGG-16 experiments, NeuCODEX consistently yields much lower latency than naive splitting. The percentage reductions (NeuCODEX vs. edge-only / vs. uncompressed) are approximately: N-Caltech 58.2\% / 97\%, Caltech 39.3\% / 96\%, CIFAR10-DVS 12.3\% / 93\%, CIFAR10 (no reduction) / ~88\% (since the bottleneck adds encoding/decoding and transmission, NeuCODEX is slightly slower in this case). These results (analogous to those for ResNet-18) show that most latency savings come from the bottleneck compression (which slashes transmission time) combined with dynamic early exit (which saves edge computation).

\subsection{Edge Energy Consumption (VGG-16)}
\begin{figure}[ht]
\centering
\includegraphics[width=\linewidth, keepaspectratio]{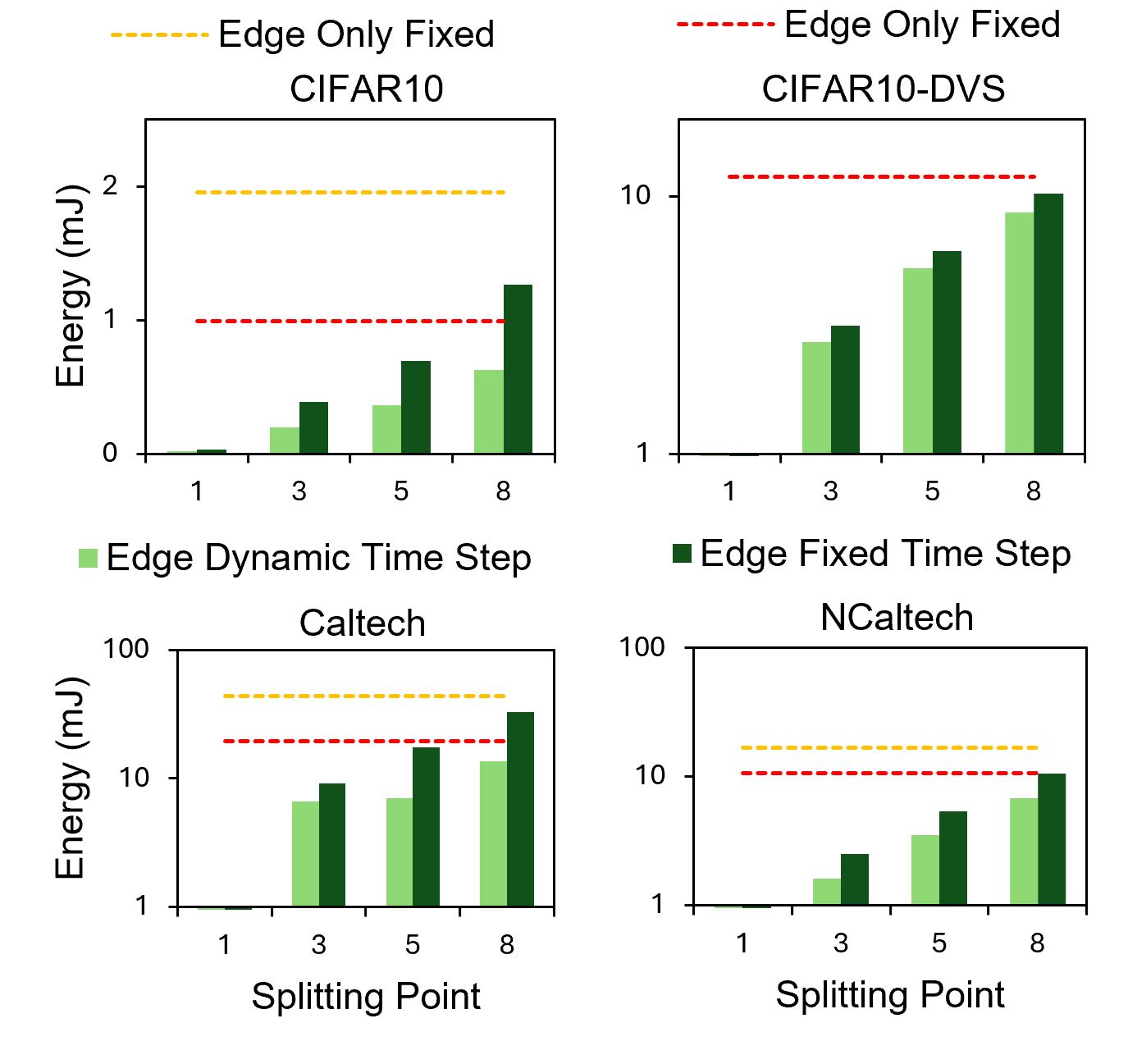}
\caption{Estimated edge-device energy consumption per inference at different network partition points for a VGG-16 SNN across various datasets. NeuCODEX greatly lowers the required energy on the edge by offloading computation after early layers and by reducing the number of timesteps processed locally.}
\label{fig:vgg_energy_consumption}
\end{figure}

Using the same Loihi-based energy model (23 pJ/SynOp), NeuCODEX at SP1 similarly reduces edge energy for VGG-16 by \textasciitilde 90–93\% across all datasets as presented in \ref{fig:vgg_energy_consumption}. For CIFAR-10, energy dropped from \textasciitilde 1.95 mJ (edge-only) to \textasciitilde 0.02 mJ (~98\% less); for CIFAR10-DVS, from \textasciitilde 13.58 mJ to \textasciitilde 0.06 mJ (\textasciitilde 99\% less); for Caltech, from \textasciitilde 43.79 mJ to \textasciitilde 0.27 mJ (\textasciitilde 99\% less); and for NCaltech, from \textasciitilde 16.80 mJ to \textasciitilde 0.04 mJ (\textasciitilde 99\% less). These dramatic savings mirror ResNet-18’s results, reflecting that only a few layers run on the edge with early exit, drastically reducing SynOps.

\vspace{12pt}
\end{document}